\def\year{2021}\relax
\documentclass[letterpaper]{article} 
\usepackage{aaai21}  
\usepackage{times}  
\usepackage{helvet} 
\usepackage{courier}  
\usepackage[hyphens]{url}  
\usepackage{graphicx} 
\usepackage{natbib}  
\usepackage{caption} 
\usepackage{amsmath}
\usepackage{graphicx}
\usepackage{amssymb}
\usepackage{multirow}
\usepackage{makecell}
\usepackage{mathrsfs}
\usepackage{threeparttable}
\usepackage[table,xcdraw]{xcolor}
\usepackage{booktabs}
\usepackage{color}
\usepackage{booktabs}
\usepackage{algorithm}
\usepackage{algorithmic}

\title{Siamese Attribute-missing Graph Auto-encoder}

\author{
Wenxuan Tu,$\footnote{Email: wenxuantu@163.com}$ Sihang Zhou, Yue Liu, Xinwang Liu\\
}
\affiliations{
    \textsuperscript{\rm 1}National University of Defense Technology\\
}

\begin{document}
\maketitle

\begin{abstract}

Graph representation learning (GRL) on attribute-missing graphs, which is a common yet challenging problem, has recently attracted considerable attention. We observe that existing literature: 1) isolates the learning of attribute and structure embedding thus fails to take full advantages of the two types of information;
2) imposes too strict distribution assumption on the latent space variables, leading to less discriminative feature representations. In this paper, based on the idea of introducing intimate information interaction between the two information sources, we propose our Siamese Attribute-missing Graph Auto-encoder (SAGA). Specifically, three strategies have been conducted. First, we entangle the attribute embedding and structure embedding by introducing a siamese network structure to share the parameters learned by both processes, which allows the network training to benefit from more abundant and diverse information. Second, we introduce a $K$-nearest neighbor (KNN) and structural constraint enhanced learning mechanism to improve the quality of latent features of the missing attributes by filtering unreliable connections. Third, we manually mask the connections on multiple adjacent matrices and force the structural information embedding sub-network to recover the true adjacent matrix, thus enforcing the resulting network to be able to selectively exploit more high-order discriminative features for data completion. Extensive experiments on six benchmark datasets demonstrate the superiority of our SAGA against the state-of-the-art methods.
\end{abstract}

\section{Introduction}
Graph representation learning (GRL), which aims to learn a graph neural network that embeds nodes to a low-dimensional latent space by preserving node attributes and graph structures, has been intensively studied and widely applied into various applications \cite{social, 2018MRL, Facial, 2020Graph}. One underlying assumption commonly adopted by these methods is that all attributes of nodes are complete. However, in practice, this assumption may not hold due to 1) the absence of particular attributes; 2) the absence of all the attributes of specific nodes. These circumstances are usually called attribute incomplete  \cite{2021GCNMF} and attribute missing \cite{2020SAT}, respectively. The existence of the above circumstances makes existing GRL methods unable to effectively handle corresponding learning problems.


To solve the first type of problem, the early methods mainly concentrate on imputation techniques or deep generative approaches for data completion, such as matrix completion via matrix factorization \cite{2017sRGCNN, 2017GCMC, 2018scf, 2019NMR} and generative adversarial learning \cite{2020GINN}. Finding that the imputation-based methods disconnect the processes of imputation and network learning, which decreases the diversity and discriminability of the learned representations, some advanced algorithms, e.g., GRAPE \cite{2020GRAPE} and GCNMF \cite{2021GCNMF}, merge the both processes of data imputation and representation learning into a united graph-based framework, where they adopt bipartite message passing strategy and Gaussian mixture model (GMM) to restore incomplete values, respectively. The above-mentioned methods could work well when handling attribute-incomplete problems. Nevertheless, they could hard to produce high-quality data completion when node attributes are entirely missing. 

Compared to the first type of methods, the second category aims to tackle a newly proposed problem, which has not been sufficiently studied in the literature and remains an open yet challenging issue. To tackle this issue, SAT \cite{2020SAT}, makes the first attempt by introducing a distribution consistency assumption between attribute and structure embedding sub-networks to guide the generation of more meaningful latent embedding.
Though demonstrating high quality of attribute restoration in various downstream tasks, SAT suffers from the following non-negligible limitations: 1) adopts two decoupled sub-networks for information extraction, thus isolates the learning of attribute and structure embedding; 2) imposes too strict distribution assumption on the latent variables, which largely decreases the discriminative capability of the learned representation; 3) it lacks a structure-attribute information filtering and refining mechanism for data completion, resulting in less robust feature representations.   


Motivated by the above observations, we propose a novel graph representation learning network for attribute-missing graphs, termed as \textbf{S}iamese \textbf{A}ttribute-missing \textbf{G}raph \textbf{A}uto-encoder (SAGA), as illustrated in Fig. 1. 
The main idea of our method is to establish a structure-attribute mutual enhanced learning strategy to 1) allow sufficient interaction between the attribute-missing matrix and the adjacent matrix for information filtering; 2) introduce a structure refinement strategy for high-quality data completion.
To facilitate the above ideas, we entangle the attribute information embedding and structural information embedding by introducing a siamese framework to share the parameters learned by the two processes.
Then, we design a dual correlation aggregating (DCA) module to filter unreliable connections by utilizing $K$-nearest neighbor (KNN) and structural constraint enhanced learning mechanism. With this mechanism, each node in the latent space could well collect and preserve the most informative information from global features. This boosts the quality of latent embedding of the restored attributes. Moreover, to enhance the quality of structural information embedding, we propose a hidden structure refining (HSR) module. 
In this module, we manually mask the connections on multiple adjacent matrices and force the structure information embedding sub-network to recover the true adjacent matrix. In this way, the network is allowed to pull closer the representations of neighbor nodes to refine their structure information, especially for that of attribute-missing nodes. This in turn enforces the resulting network to be able to selectively exploit more discriminative features from hidden high-order attributes for data completion. The main contributions of this paper are listed as follows:

\begin{itemize}
\item We propose a novel graph representation learning framework termed as \textbf{S}iamese \textbf{A}ttribute-missing \textbf{G}raph \textbf{A}uto-encoder (SAGA) to handle attribute-missing graphs, which allows us to elegantly achieve powerful data completion without any prior assumption.
 
\item By promoting the learning processes of attribute and structure information to sufficiently interact with each other via DCA and HSR modules, we can take full advantages of two-source information to filter unreliable connections as well as preserve more informative information in the latent space. In this way, the discriminative capacity of resultant latent embedding is improved.

\item Extensive experimental results on six benchmark datasets demonstrate that our proposed method is highly competitive and consistently outperforms the state-of-the-art ones with a preferable margin.
\end{itemize}

\section{Related Work}
\noindent{\textbf{Siamese Networks}}. Siamese network is a kind of network that contains more than one identical weight-sharing sub-networks, which can naturally introduce inductive biases for invariance modeling \cite{1993}. It has wide applications including video super-resolution \cite{2019tsa}, medical object detection \cite{2020aa}, visual tracking \cite{2021STM}, etc. Inspired by its success in various visual tasks, researchers have successfully introduced siamese networks into the field of graph learning \cite{2021bem, 2021MERIT}. However, it has not been extended to handle incomplete or missing graphs. 

\noindent{\textbf{Graph Representation Learning}}. 
Early graph representation learning (GRL) methods learn node embeddings by utilizing probability models on the generated random walk paths on graphs
\cite{2014DeepWalk, 2016node2vec}. However, these methods overly emphasize the structural information while ignore the important attribute information.
Thanks to the development of graph neural networks (GNNs), GNN-based GRL methods that jointly exploit graph structure information and node attribute information in spectral or spatial domain have been widely studied in recent years. These methods can be roughly categorized into auto-encoder-based methods \cite{2016Variational, 2019Adversarial, Pan2019Learning, Bo2020Structural, 2021DFCN} and contrastive learning-based methods \cite{2019Deep, 2020Contrastive, 2020Graph, 2020GraphContrastive, 2021GCA}. 
One underlying assumption commonly adopted by current GRL methods is that all node attributes are complete. However, in practice, the absent data makes it difficult to utilize the existing methods for satisfactory performance.

\noindent{\textbf{Graph Deep Learning with Absent Data}}. 
To handle incomplete graph data, one popular way commonly adopted by existing algorithms is data imputation technique, such as matrix completion and generative adversarial learning. For matrix completion, sRGCNN \cite{2017sRGCNN}, GC-MC \cite{2017GCMC}, and NMTR \cite{2019NMR} formulate the user-item rating matrix, users (or items), and the observed ratings as bipartite graph, nodes, and links, respectively. Then they apply a graph auto-encoder (GAE) to predict the absent linkages between node pairs.
By introducing an additional adversarial loss, GINN \cite{2020GINN} trains a graph denoising auto-encoder to build intermediate representations of all nodes with the help of a pre-processing graph for data imputation. 
To further improve the quality of attribute restoration, recent efforts combine the processes of data imputation and representation learning into a united framework. For instance, GRAPE \cite{2020GRAPE} formulates the feature imputation as an edge-level prediction task on the graph. 
After that, GCNMF \cite{2021GCNMF} estimates the incomplete variables by enforcing them to follow the Gaussian mixture distribution using a Gaussian mixture model (GMM).

Although these methods can be competent to effectively handle attribute-incomplete problem, they fail to perform preferably on datasets with missing attributes, i.e., the entire attributes of specific nodes are missing. 
More recently, an advanced algorithm SAT is proposed to learn on attribute-missing graphs. It adopts two decoupled sub-networks to process node attributes and graph structures, and then utilize structure information to conduct data completion under the guidance of a shared-latent space assumption \cite{2020SAT}. Despite its success, SAT not only isolates the learning processes of attribute and structure embeddings but also heavily relies on a prior distribution assumption for latent representation learning. Thus it fails to take full advantages of both types information, leading to less discriminative latent representations. In contrast, our method allows both learned features to sufficiently interact with each other via a structure-attribute mutual enhanced learning strategy. As a results, the learned latent embedding has the potential to be more discriminative for data completion.

\begin{figure}[!t]
\centering
\includegraphics[width=3.3in]{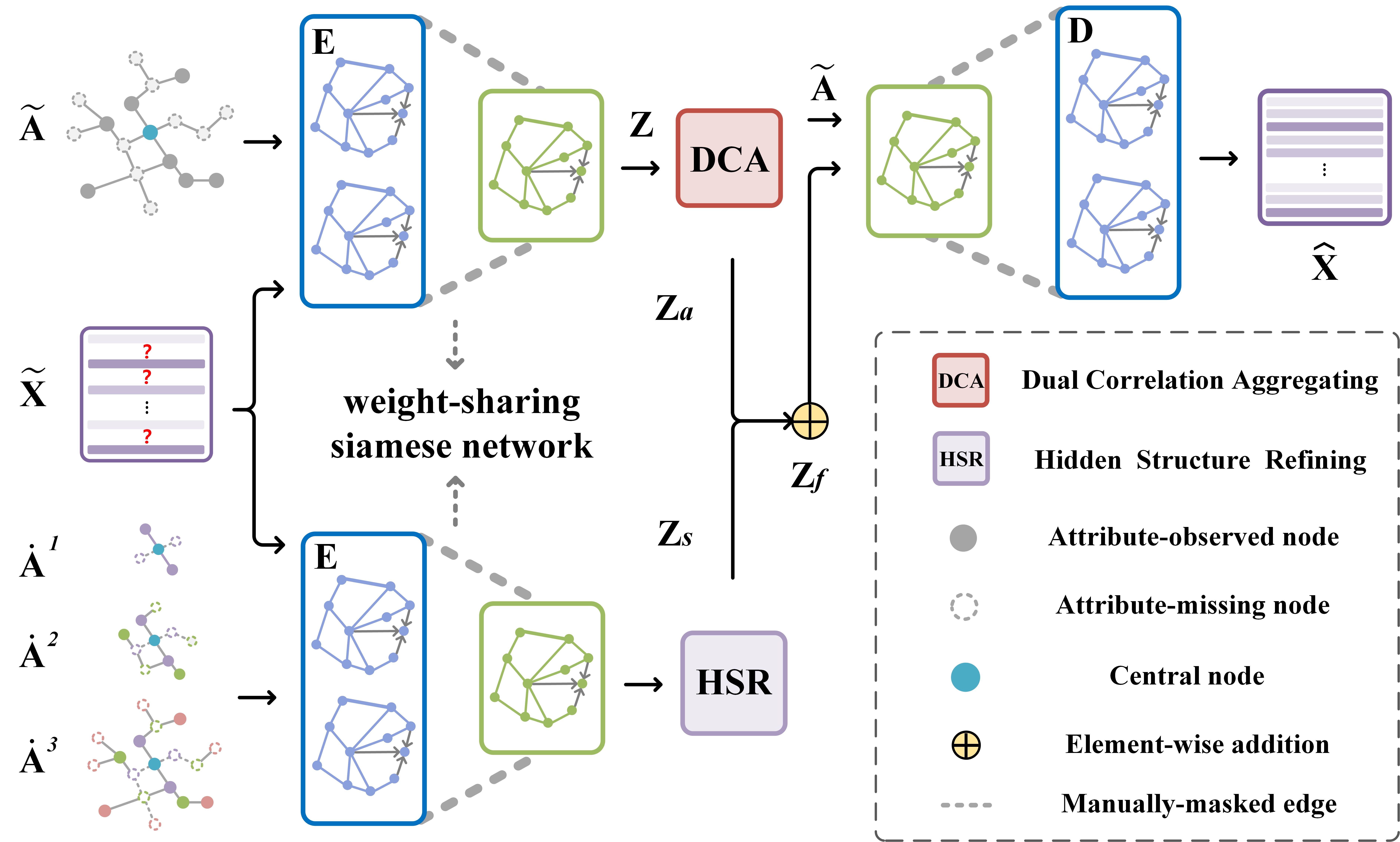}
\caption{Illustration of SAGA. Our siamese architecture consists of two branches where the attributes and the affinity matrices are closely entangled. Specially, in the upper branch, the DCA module improves the quality of missing attribute latent feature learning by introducing an unreliable similarity filtering mechanism.
While in the bottom branch, the HSR module adopts a multi-order edge recover strategy to make the network able to exploit intrinsic data structures for information recovery.}
\label{1}
\end{figure}

\section{The Proposed Method}
Our SAGA mainly consists of two components, i.e., a dual correlation aggregating (DCA) module and a hidden structure refining (HSR) module. 
In the following sections, we will provide details on notations, two carefully-designed components, and the optimization target, respectively.

\subsection{Notations}
Given an undirected graph $\mathcal{G}=\{\mathcal{V, E}\}$ with $C$ classes, where $\mathcal{V} = \{v_{1}, v_{2}, \dots, v_{N}\}$, $\mathcal{E}$, and $N$ are the node set, edge set, and the number of nodes, respectively. In classic graph learning, a graph is usually characterized by its attribute matrix $\mathbf{X} \in \mathbb{R}^{N \times D}$ and normalized adjacency matrix $\widetilde{\mathbf{A}} \in \mathbb{R}^{N \times N}$, where $D$ is the node attribute dimension \cite{2017GCN}. 
Specially, in attribute-missing graph learning, with the existence of missing attributes, we further define $\mathcal{V}^{o} = \{v_{1}^{o}, v_{2}^{o}, \dots, v_{N_{o}}^{o}\}$ and $\mathcal{V}^{m} = \{v_{1}^{m}, v_{2}^{m}, \dots, v_{N_{m}}^{m}\}$ to be the set of attribute-observed nodes and the set of attribute-missing nodes, respectively.
Accordingly, $\mathcal{V}$ = $\mathcal{V}^{o}$ $\cup$ $\mathcal{V}^{m}$, $\mathcal{V}^{o}$ $\cup$ $\mathcal{V}^{m}$ = $\varnothing$ and $N$ = $N_{o}$ + $N_{m}$. 
In this circumstance, the missing attributes in $\mathbf{X}$ is firstly filled with all zeros and the resulting matrix is denoted as $\widetilde{\mathbf{X}} \in \mathbb{R}^{N \times D}$.
To introduce high-order adjacent information, we construct a series of adjacent matrices of different orders $\mathcal{A} = \{\mathbf{A}^{\textit{1}}, \mathbf{A}^{\textit{2}}, \dots, \mathbf{A}^{H}\}$, where $\mathbf{A}^{h} \in \mathbb{R}^{N \times N}$, $H$ is the number of orders. Specially, $\mathbf{A}^{h} = \mathbf{A}^{\textit{1}}\mathbf{A}^{(h-1)}$, where we denote $\mathbf{A}^{\textit{0}}$ as identity matrix $\mathbf{I} \in \mathbb{R}^{N \times N}$, 1 $\leq h \leq H$. 
During the training, we manually mask partial connections on multi-order adjacent matrices to boost the network learning, thus $\mathcal{A}$ is redefined as $\mathcal{\dot{A}} = \{\mathbf{\dot{A}}^{\textit{1}}, \mathbf{\dot{A}}^{\textit{2}}, \dots, \mathbf{\dot{A}}^{H}\}$, where $\mathbf{\dot{A}}^{h} \in \mathbb{R}^{N \times N}$. Table \ref{I} summarizes the commonly used notations.

\subsection{Structure-attribute Mutual Enhancement}
\begin{figure}[!t]
\centering
\includegraphics[width=3.2in]{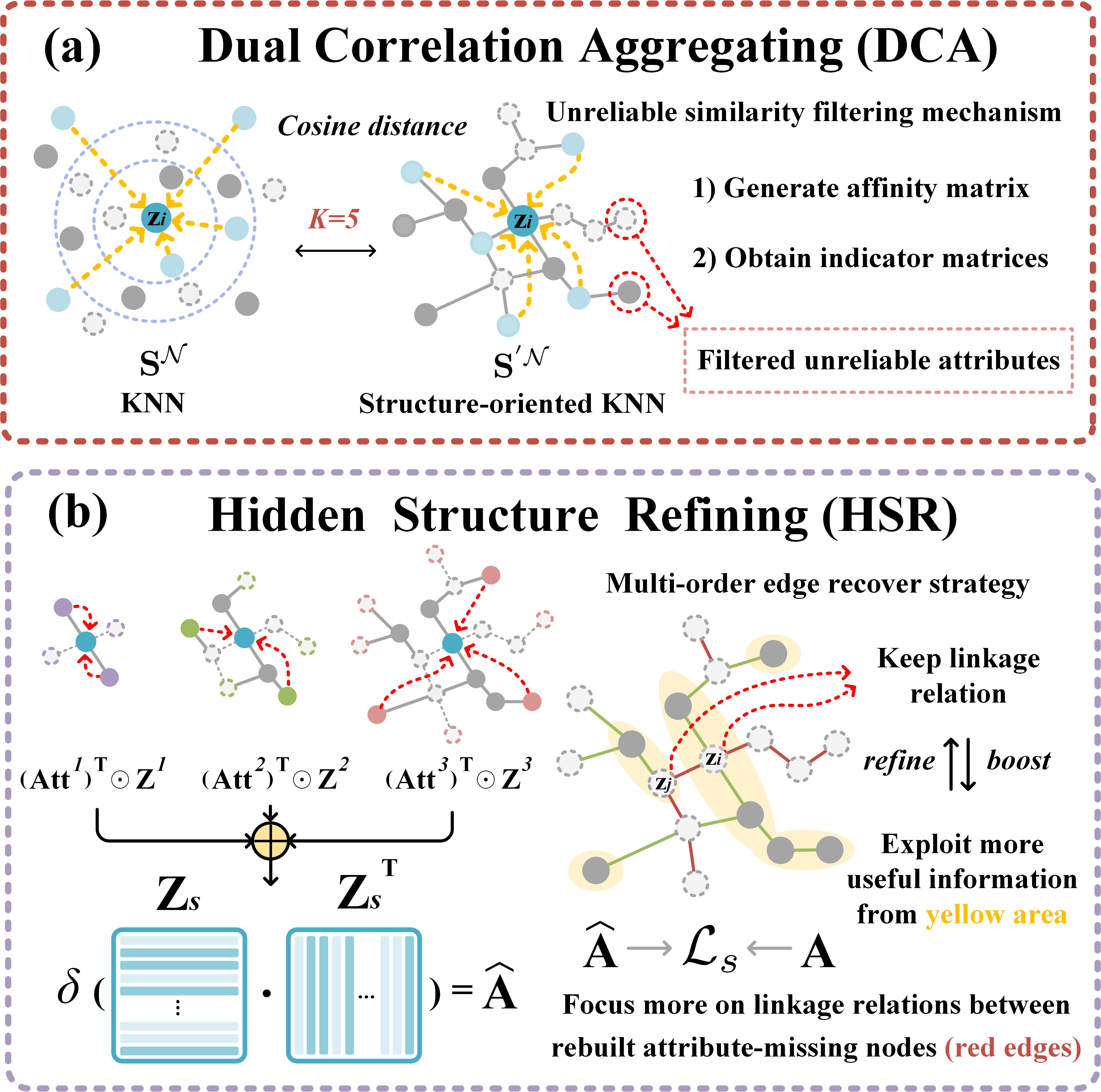}
\caption{Illustration of DCA module and HSR module.}
\label{2}
\end{figure}

In this part, we introduce the two proposed components, i.e., the dual correlation aggregating (DCA) module and the hidden structure refining (HSR) module in detail. Both modules are illustrated in Fig. \ref{2}. The information of the shared siamese network will be introduced during the introduction of the two parts. 
\begin{table}[!t]
\centering

\footnotesize
\begin{tabular}{l|l}\hline
\hline
\multicolumn{1}{c|}{Notations} & \multicolumn{1}{c}{Meaning}   \\\hline
$\mathbf{X} \in \mathbb{R}^{N \times D}$    & Original attribute matrix   \\
$\mathbf{A} \in \mathbb{R}^{N \times N}$    & Original adjacency matrix   \\
$\widetilde{\mathbf{X}} \in \mathbb{R}^{N \times D}$    & Zero-filled attribute matrix   \\
$\widetilde{\mathbf{A}} \in \mathbb{R}^{N \times N}$    & Normalized adjacency matrix   \\
$\mathbf{A}^{h} \in \mathbb{R}^{N \times N}$  & Normalized $h$-order adjacency matrix  \\
$\mathbf{\dot{A}}^{h} \in \mathbb{R}^{N \times N}$  & Edge-masked $h$-order adjacency matrix  \\
$\mathbf{Z} \in \mathbb{R}^{N \times d}$  & Latent embedding matrix    \\
$\mathbf{Z}_{a} \in \mathbb{R}^{N \times d}$  & Attribute-enhanced latent embedding matrix    \\
$\mathbf{Z}^{h} \in \mathbb{R}^{N \times d}$  & $h$-th path latent embedding matrix    \\
$\mathbf{Z}_{s} \in \mathbb{R}^{N \times d}$  & Structure-enhanced latent embedding matrix    \\
$\mathbf{Z}_{f} \in \mathbb{R}^{N \times d}$  & Fused latent embedding matrix    \\
$\widehat{\mathbf{A}} \in \mathbb{R}^{N \times N}$  & Rebuilt adjacency matrix    \\
$\widehat{\mathbf{X}} \in \mathbb{R}^{N \times D}$  & Rebuilt attribute matrix    \\
\hline\hline
\end{tabular}
\caption{Summary of notations}
\label{I}
\end{table}

\noindent{\textbf{Dual Correlation Aggregating}}.
For given inputs $\widetilde{\mathbf{X}}$ and $\widetilde{\mathbf{A}}$, a siamese graph encoder $E$ conducts the following layer-wise propagation in hidden layers as:
\begin{equation} \label{eq:1}
\mathbf{Z}^{(l)} = \sigma(\widetilde{\mathbf{A}}\mathbf{Z}^{(l-1)}\mathbf{W}^{(l)}),
\end{equation}
where $\mathbf{W}^{(l)}$ denotes the learnable parameters of the \textit{l}-th encoder layer. $\sigma$ is a non-linear activation function, e.g., ReLU. We denote $\mathbf{Z}^{0}$ as zero-filled attribute matrix $\widetilde{\mathbf{X}}$. 
As we can see, the GCN conducts neighborhood aggregation operation in each layer, it can be viewed as neighbor imputation. Thus, the missing attributes are gradually imputed during the calculation of the network and become complete in the latent embedding matrix $\mathbf{Z}$. 
Although the neighborhood aggregation could provide primary imputation, the imputed value could be noisy and with little discriminative capability. To this end, we further refine it by performing accurate global information enhancement to introduce reliable global information with few layers. Specially, we introduce an extra operation after the encoder as follow:
\begin{equation} \label{eq:4}
\mathbf{Z}_{a} =\alpha \mathbf{S}^{\mathcal{N}} \mathbf{Z} +(1-\alpha) \mathbf{S}^{'\mathcal{N}} \mathbf{Z},
\end{equation}
where $\alpha$ is the learnable weighting coefficient and we set $\alpha$ = 0.5 for initialization. $\mathbf{S}^{\mathcal{N}}, \mathbf{S}^{'\mathcal{N}} \in \mathbb{R}^{N \times N}$ are two refined global similarity matrices (i.e., indicator matrices) which are constructed in two different manners. As shown in Fig. \ref{2}(a), to construct $\mathbf{S}^{\mathcal{N}}$ and $\mathbf{S}^{'\mathcal{N}}$, we first generate the affinity matrix according to the latent embedding matrix as follow:
\begin{equation} \label{eq:2}
\mathbf{S}_{ij} = \frac{{\mathbf{z}_{i}\mathbf{z}_{j}^{\mathbf{T}}}}{{\Vert \mathbf{z}_{i} \Vert \Vert \mathbf{z}_{j} \Vert}}, \,\,\,\,  \forall\,\,i, j \in [1, N].
\end{equation}
Here $\mathbf{z}_{i}$ indicates the latent embedding of the $i$-th sample.
To improve the reliability of $\mathbf{S}$, two mechanisms are introduced.
 
To construct $\mathbf{S}^{\mathcal{N}}$, we adopt a $\textit{K}$-nearest neighbors (KNN) strategy to filter the less confident similarity estimation in $\mathbf{S}$. Specifically, only the largest similarity values of each of the samples are kept while other values are assigned as $0$.
To construct $\mathbf{S}^{'\mathcal{N}}$, we first collect all the neighbors in the $1$- to $P$- order affinity matrix, then filter the non-neighbor elements in $\mathbf{S}$ by setting them to $-1$. Finally, we further conduct a KNN strategy to get the final refined similarity matrix.

With the refined similarity matrices, we update the latent embedding with Eq. (2) for reliable global information aggregation, so as to improve the quality of missing attribute imputation and the discriminative capability of the latent features simultaneously.

\begin{algorithm}[!htbp]
{\caption{Training procedure of SAGA}\label{Algorithm-proposed}
\small
\begin{algorithmic}[1]
\REQUIRE Zero-filled attribute matrix $\widetilde{\mathbf{X}}$; Normalized adjacency matrix $\widetilde{\mathbf{A}}$; A series of edge-masked adjacent matrices of different orders $\mathcal{\dot{A}}$; Iteration number $\textit{I}$; Hyper-parameters $P$, $K$, $\gamma$, $\lambda$.
\ENSURE Rebuilt attribute matrix $\widehat{\mathbf{X}}$.
\STATE Initialize the model parameters $\theta$ with an Xavier initialization;
\FOR {$\textit{i} = 1$ to $\textit{I}$}
\STATE Utilize $E$ to encode $\mathbf{Z}$ by Eq. (1);
\STATE Construct $\mathbf{S}^{\mathcal{N}}$ and $\mathbf{S}^{'\mathcal{N}}$ using KNN by Eq. (3);
\STATE Calculate $\mathbf{Z}_{a}$ by Eq. (2);
\STATE Utilize $E$ to encode \{$\mathbf{Z}^{\textit{1}}$, $\mathbf{Z}^{\textit{2}}$, $\dots$, $\mathbf{Z}^{H}$\} by Eq. (4);
\STATE Calculate $\mathbf{Z}_{s}$ by Eq. (5) and Eq. (6);
\STATE Rebuilt $\widehat{\mathbf{A}}$ based on $\mathbf{Z}_{s}$ by Eq. (7);
\STATE Calculate $\mathcal{L}_{s}$ by Eq. (8), Eq. (9), and Eq. (10);
\STATE Fuse $\mathbf{Z}_{f}$ using $\mathbf{Z}_{a}$ and $\mathbf{Z}_{s}$ by Eq. (11);
\STATE Utilize $D$ to decode $\mathbf{Z}_{f}$ and output $\widehat{\mathbf{X}}$ by Eq. (12);
\STATE Optimize the network with Adam by minimizing Eq. (13);
\ENDFOR
\RETURN $\widehat{\mathbf{X}}$
\end{algorithmic}}
\end{algorithm}

\noindent{\textbf{Hidden Structure Refining}}.
In addition to the information filtering mechanism that guarantees the quality of attribute restoration, the structure information of attribute-missing nodes has not been fully considered yet, which acts as a strong constraint to their linkage relations in the graph. Fig. \ref{2}(b) illustrates the design of HSR module, which consists of two schemes, i.e., the multi-order observed attributes fusion and the edge recovery. 

Likewise, by transferring given inputs, i.e., $\widetilde{\mathbf{X}}$ and $\mathcal{\dot{A}} = \{\mathbf{\dot{A}}^{\textit{1}}, \mathbf{\dot{A}}^{\textit{2}}, \dots, \mathbf{\dot{A}}^{H}\}$, into a weight-sharing graph encoder $E$, we model the $h$-th path latent representations $\mathbf{Z}^{h(l)}$ as: 
\begin{equation} \label{eq:5}
\mathbf{Z}^{h(l)} = \sigma(\mathbf{\dot{A}}^{h}\mathbf{Z}^{h(l-1)}\mathbf{W}^{(l)}),
\end{equation}
where we consider $1$- to $3$- order neighbors, i.e., $H$=3, $h \in [1,H]$. $\mathbf{Z}^{h(0)}$ and $\mathbf{Z}^{h}$ are denoted as $\widetilde{\mathbf{X}}$ and the $h$-th path latent embedding matrix where each node aggregates $h$-order observed attributes, respectively. Then we transform these embedding matrices though a nonlinear transformation (e.g., one-layer MPL) to estimate the importance of each path. For node $n$, where its embedding in $\mathbf{Z}^{h}$ is $\mathbf{z}_{n}^{h} \in \mathbb{R}^{1 \times d}$, an normalized attention weight $a^{h}_{n}$ using softmax function is formulated as follows:

\begin{equation} \label{eq:6}
a^{h}_{n} = \frac{e^{(\mathbf{W}^{h}(\mathbf{z}_{n}^{h})^{\mathbf{T}}+\mathbf{b}^{h})}}{\sum_{h=1}^H e^{(\mathbf{W}^{h}(\mathbf{z}_{n}^{h})^{\mathbf{T}}+\mathbf{b}^{h})}},
\end{equation}
where $\mathbf{W}^{h} \in \mathbb{R}^{d \times 1}$ denotes the learnable attention parameters and $\mathbf{b}^{h} \in \mathbb{R}^{d \times 1}$ denotes the bias vector of $h$-th path. Larger $a^{h}_{n}$ illustrates that the $h$-order observed neighbors could provide more informative information for node $n$. Next, we combine these latent embedding matrices with attention weights:
\begin{equation} \label{eq:7}
\mathbf{Z}_{s} = \sum_{h=1}^H (\mathbf{Att}^{h})^{\mathbf{T}} \odot \mathbf{Z}^{h},
\end{equation}
where $\odot$ means matrix product, $\mathbf{Att}^{h} \in \mathbb{R}^{d \times N}$ is denoted as [$\mathbf{a}^{h}_{1}$, $\mathbf{a}^{h}_{2}$, \dots, $\mathbf{a}^{h}_{N}$] and $\mathbf{a}^{h}_{n} \in \mathbb{R}^{d \times 1}$ is an attention vector that repeats $a^{h}_{n}$ with $d$ times. After that, $\mathbf{Z}_{s}$ is decoded into $\widehat{\mathbf{A}}$ through matmul product with an activation function:
\begin{equation} \label{eq:8}
\widehat{\mathbf{A}} = \textit{Sigmoid} (\mathbf{Z}_{s} \mathbf{Z}_{s}^{\mathbf{T}}).  
\end{equation}
According to Eq. (7), we can model the linkage relation between node $i$ and node $j$ by minimizing:
\begin{equation} \label{eq:9}
l_{ij} = -[\mathbf{A}_{ij} \ln \widehat{\mathbf{A}}_{ij} + (1-\mathbf{A}_{ij}) \ln (1-\widehat{\mathbf{A}}_{ij})],  
\end{equation}
where there exists a linkage between node $i$ and node $j$ in original graph if $\mathbf{A}_{ij}$ = 1, otherwise 0.
To enable the network to focus more on attribute-missing nodes to exploit their intrinsic structures, we introduce a pre-defined hyper-parameter $\gamma$ to balance the edge recovery processes of two types of linkage relations, i.e., the manually-masked part and the manually-preserved part, corresponding to the red and green lines in Fig. \ref{2}(b). Eq. (8) can be reformulated as:
\begin{equation}  \label{eq:10}
\mathcal{L}_{ij} = \left\{
\begin{array}{ll}
\gamma l_{ij}, & \text{$v_{i}$, $v_{j}$ $\in \mathcal{V}^{m}$}\\
\\
 l_{ij}, & \text{otherwise}\\
\end{array}.
\right.
\end{equation}
This is very different from the structure preservation fashion of missing nodes as in SAT \cite{2020SAT} that we manually mask the edge between attribute-missing nodes on multiple graphs and enforce the network to focus more on these linkage relations. We summarize the merits of the fashion of our edge recovery: 1) more naturally evaluates the overall quality of restored attributes via structure refinement; 2) in turn enforces the resulting network to be able to selectively exploit more high-order discriminative information for data completion. Finally, the average structure reconstruction loss of all node pairs can be written as:
\begin{equation} \label{eq:11}
\mathcal{L}_{s} = \frac{1}{N^{2}}\sum_{i=1}^N \sum_{j=1}^N\mathcal{L}_{ij}.  
\end{equation}

\noindent{\textbf{Information Aggregation and Decoding}}. After obtaining the attribute-enhanced latent embedding matrix $\mathbf{Z}_{a}$ and the structure-enhanced latent embedding matrix $\mathbf{Z}_{s}$ from DCA module and HSR module, we combine both with a learnable weighting coefficient $\beta$, where $\beta$ is initialized as 0.5. Then we directly feed the fused latent embedding matrix $\mathbf{Z}_{f}$ with $\widetilde{\mathbf{A}}$ into a graph decoder $D$. This process is formulated as:
\begin{equation} \label{eq:12}
\mathbf{Z}_{f} = \beta \mathbf{Z}_{a} + (1-\beta)\mathbf{Z}_{s},
\end{equation}

\begin{equation} \label{eq:13}
\mathbf{Z}^{'(l)} = \sigma(\widetilde{\mathbf{A}}\mathbf{Z}^{'(l-1)}\mathbf{W}^{'(l)}),
\end{equation}
where $\mathbf{W}^{'(l)}$ denotes the learnable parameters of the \textit{l}-th decoder layer. $\mathbf{Z}^{'(0)}$ and $\widehat{\mathbf{X}} \in \mathbb{R}^{N \times D}$ are denoted as $\mathbf{Z}_{f}$ and the rebuilt attribute matrix, respectively. 

\subsection{Joint Loss and Optimization}
The overall learning objective consists of two parts, i.e., the attribute reconstruction loss of SAGA, and the structure reconstruction loss which is correlated with HSR module:
\begin{equation} \label{eq:14}
\mathcal{L}_{total} = \lambda \mathcal{L}_{a} + \mathcal{L}_{s}.
\end{equation}
In Eq.\eqref{eq:14}, $\mathcal{L}_{a}$ denotes the mean square error (MSE) between the observed parts of $\mathbf{\widetilde{X}}$ and $\mathbf{\widehat{X}}$. $\lambda$ is a pre-defined hyper-parameter which balances the importance of both reconstruction processes. The detailed learning procedure of the proposed SAGA is shown in Algorithm 1.

\section{Experiments}
\subsection{Experimental Setup}
\subsubsection{Benchmark Datasets}
We evaluate the proposed SAGA on six benchmark datasets, including four datasets with categorial attributes, i.e., Cora \cite{2000Cora}, Citeseer \cite{2008Cite}, Amazon-Computer and Amazon-Photo \cite{2018amaz}, and two datasets with real-valued attributes, i.e., Pubmed \cite{2008Cite} and Coauthor-CS \cite{2012Cocs}.
Table \ref{II} summarizes the brief information of these datasets.


\subsubsection{Parameters Setting}
We follow the same data splits as in SAT \cite{2020SAT} on all benchmark datasets, including the split of attribute-observed/-missing nodes and the spilt of train/test sets. For GINN \cite{2020GINN} and GCNMF \cite{2021GCNMF}, we set the hyper-parameters of both methods by following their original papers. For other compared methods, we report the results listed in the paper of SAT directly. For our SAGA, we adopt 4-layer GCNs as our backbone and train it with Adam optimizer, where the learning rate is set to 1e-3. We first conduct profiling task using Recall@K and NDCG@K as metrics to evaluate the quality of restored attributes by training SAGA for at least 500 iterations until convergence. Then we utilize a GCN-based classifier to perform node classification task over the rebuilt graph with five-fold validation in 10 times, and report the averages evaluated by accuracy (ACC). We adopt an early stop strategy when the loss value comes to a plateau to avoid over-fitting. According to the results of hyper-parameter analysis, we set the hyper-parameters $P$, $K$, and $\gamma$ as 5, and fix $\lambda$ to 10. More implementation details are presented in Appendix A.

\begin{table}[!t]
\centering

\footnotesize
\begin{tabular}{c|c|c|c|c}\hline
\hline
Dataset    & Nodes   & Edges    & Dimension  & Classes\\\hline
Cora       & 2708    & 10556    & 1433       & 7       \\
Citeseer   & 3327    & 9228     & 3703       & 6       \\
Amazon-C   & 13752   & 574418   & 767        & 10      \\
Amazon-P   & 7650    & 287326   & 745        & 8      \\
Pubmed     & 19717   & 88651    & 500        & 3       \\
Coauthor-CS    & 18333  & 327576 & 6805      & 15\\\hline\hline
\end{tabular}
\caption{Dataset summary}
\label{II}
\end{table}

\begin{table*}[!t]
\centering
\label{tab:my-table}
\scriptsize
\begin{tabular}{c|c|ccccccccc|c}
\hline\hline
Dataset               & Metric & NeighAggre & VAE    & GCN   & GraphSage  & GAT   & Hers & GraphRNA & ARWMF & SAT & Ours  
\\\hline
\multirow{6}{*}{Cora} & Recall@10    & 0.0906  & 0.0887  & 0.1271  & 0.1284  & 0.1350  & 0.1226  & 0.1395  & 0.1291  & {\color{blue}0.1508}  & {\color{red}0.1734}     \\
                      & Recall@20    & 0.1413  & 0.1228  & 0.1772  & 0.1784  & 0.1812  & 0.1723  & 0.2043  & 0.1813  & {\color{blue}0.2182}  & {\color{red}0.2410}     \\
                      & Recall@50    & 0.1961  & 0.2116  & 0.2962  & 0.2972  & 0.2972  & 0.2799  & 0.3142  & 0.2960  & {\color{blue}0.3429}  & {\color{red}0.3620}     \\
                      & NDCG@10      & 0.1217  & 0.1224  & 0.1736  & 0.1768  & 0.1791  & 0.1694  & 0.1934  & 0.1824  & {\color{blue}0.2112}  & {\color{red}0.2382}     \\
                      & NDCG@20      & 0.1548  & 0.1452  & 0.2076  & 0.2102  & 0.2099  & 0.2031  & 0.2362  & 0.2182  & {\color{blue}0.2546}  & {\color{red}0.2831}     \\
                      & NDCG@50      & 0.1850  & 0.1924  & 0.2702  & 0.2728  & 0.2711  & 0.2596  & 0.2938  & 0.2776  & {\color{blue}0.3212}  & {\color{red}0.3472}     \\\hline
\multirow{6}{*}{Citeseer} 
                      & Recall@10    & 0.0511 & 0.0382 & 0.0620 & 0.0612 & 0.0561 & 0.0576 & {\color{blue}0.0777} & 0.0552 & 0.0764 & {\color{red}0.0945} \\
                      & Recall@20    & 0.0908 & 0.0668 & 0.1097 & 0.1097 & 0.1012 & 0.1025 & 0.1272 & 0.1015 & {\color{blue}0.1280} & {\color{red}0.1535} \\
                      & Recall@50    & 0.1501 & 0.1296 & 0.2052 & 0.2058 & 0.1957 & 0.1973 & 0.2271 & 0.1952 & {\color{blue}0.2377} & {\color{red}0.2674} \\
                      & NDCG@10      & 0.8023 & 0.0601 & 0.1026 & 0.1003 & 0.0878 & 0.0904 & 0.1291 & 0.0859 & {\color{blue}0.1298} & {\color{red}0.1615} \\
                      & NDCG@20      & 0.1155 & 0.0839 & 0.1423 & 0.1393 & 0.1253 & 0.1279 & 0.1703 & 0.1245 & {\color{blue}0.1729} & {\color{red}0.2107} \\
                      & NDCG@50      & 0.1560 & 0.1251 & 0.2049 & 0.2034 & 0.1872 & 0.1900 & 0.2358 & 0.1858 & {\color{blue}0.2447} & {\color{red}0.2858} \\\hline
\multirow{6}{*}{Amazon-C}
                      & Recall@10    & 0.0321 & 0.0255 & 0.0273 & 0.0269 & 0.0271 & 0.0273 & 0.0386 & 0.0280 & {\color{blue}0.0391} & {\color{red}0.0445} \\
                      & Recall@20    & 0.0593 & 0.0502 & 0.0533 & 0.0528 & 0.0530 & 0.0525 & 0.0690 & 0.0544 & {\color{blue}0.0703} & {\color{red}0.0787} \\
                      & Recall@50    & 0.1306 & 0.1196 & 0.1275 & 0.1278 & 0.1278 & 0.1273 & 0.1465 & 0.1289 & {\color{blue}0.1514} & {\color{red}0.1651} \\
                      & NDCG@10      & 0.0788 & 0.0632 & 0.0671 & 0.0664 & 0.0673 & 0.0676 & 0.0931 & 0.0694 & {\color{blue}0.0963} & {\color{red}0.1088} \\
                      & NDCG@20      & 0.1156 & 0.0970 & 0.1027 & 0.1020 & 0.1028 & 0.1025 & 0.1333 & 0.1053 & {\color{blue}0.1379} & {\color{red}0.1544} \\
                      & NDCG@50      & 0.1923 & 0.1721 & 0.1824 & 0.1822 & 0.1830 & 0.1825 & 0.2155 & 0.1851 & {\color{blue}0.2243} & {\color{red}0.2463} \\\hline
\multirow{6}{*}{Amazon-P}  
                      & Recall@10    & 0.0329 & 0.0276 & 0.0294 & 0.0295 & 0.0294 & 0.0292 & 0.0390 & 0.0294 & {\color{blue}0.0410} & {\color{red}0.0455} \\
                      & Recall@20    & 0.0616 & 0.0538 & 0.0573 & 0.0562 & 0.0573 & 0.0574 & 0.0703 & 0.0568 & {\color{blue}0.0743} & {\color{red}0.0792} \\
                      & Recall@50    & 0.1361 & 0.1279 & 0.1324 & 0.1322 & 0.1324 & 0.1324 & 0.1328 & {\color{blue}0.1508} & 0.1327 & {\color{red}0.1633} \\
                      & NDCG@10      & 0.0813 & 0.0675 & 0.0705 & 0.0712 & 0.0705 & 0.0714 & 0.0959 & 0.0727 & {\color{blue}0.1006} & {\color{red}0.1108} \\
                      & NDCG@20      & 0.1196 & 0.1031 & 0.1082 & 0.1079 & 0.1083 & 0.1094 & 0.1377 & 0.1098 & {\color{blue}0.1450} & {\color{red}0.1557} \\
                      & NDCG@50      & 0.1998 & 0.1830 & 0.1893 & 0.1896 & 0.1892 & 0.1906 & 0.2232 & 0.1915 & {\color{blue}0.2395} & {\color{red}0.2451} \\\hline\hline
\end{tabular}
\caption{Profiling performance.  
The {\color{red}red} and {\color{blue}blue} values indicate the best and the runner-up results, respectively.}
\label{III}

\end{table*}

\begin{table*}[!t]
\centering
\label{tab:my-table}
\scriptsize
\begin{tabular}{c|ccccccccccc|c}
\hline\hline
Dataset               & NeighAggre & VAE    & GCN   & GraphSage  & GAT   & Hers & GraphRNA & ARWMF  & GINN & GCNMF & SAT & Ours  
\\\hline
\multirow{1}{*}{Cora}
                        & 0.6494  & 0.3011  & 0.4387  & 0.5779  & 0.4525  & 0.3405  & 0.8198  & 0.8025  & 0.6758   & 0.7030& {\color{blue}0.8327} &  {\color{red}0.8513}  \\
\multirow{1}{*}{Citeseer} 

                           & 0.5413 & 0.2663 & 0.4079 & 0.4278 & 0.2688 & 0.3229 & 0.6394 & 0.2764  & 0.5532 & 0.6340 & {\color{blue}0.6599} &  {\color{red}0.6925} \\
\multirow{1}{*}{Amazon-C}
                           & 0.8715 & 0.4023 & 0.3974 & 0.4019 & 0.4034 & 0.4025 & {\color{blue}0.8650} & 0.7400  & 0.8172 &0.7643& 0.8519 & {\color{red}0.8865}  \\
\multirow{1}{*}{Amazon-P}  
                          & 0.9010 & 0.3781 & 0.3656 & 0.3784 & 0.3789 & 0.3794 & {\color{blue}0.9207} & 0.6146  & 0.8777  &0.8779& 0.9163 & {\color{red}0.9236} \\
\multirow{1}{*}{Pubmed} 
                           & 0.6564 & 0.4007 & 0.4203 & 0.4200 & 0.4196 & 0.4205 & {\color{red}0.8172} & {\color{blue}0.8089} & 0.5428 & 0.6200 & 0.7537&  0.8055 \\
\multirow{1}{*}{Coauthor-CS}
                         & 0.8031 & 0.2335 & 0.2180 & 0.2335 & 0.2334 & 0.2334 & {\color{blue}0.8851} & 0.8347  & 0.7974 & 0.8753 & 0.8576 & {\color{red}0.8890}  \\
                          \hline\hline
                      
\end{tabular}
\caption{Node classification performance. The {\color{red}red} and {\color{blue}blue} values indicate the best and the runner-up results, respectively.}
\label{IV}
\end{table*}

\subsection{Baselines and Comparison Results}
In the following, we compare our SAGA with 11 related methods to illustrate its effectiveness. Among these baselines, NeighAggre \cite{2008Navigating} is the representative one of classical profiling algorithms. VAE \cite{2014VAE} is a well-known auto-encoder based generative method. GCN \cite{2017GCN}, GraphSage \cite{2017Graphsage}, and GAT \cite{2018GAN} are typical graph convolutional network (GCN)-based methods, where the node representations are embedded with structure information by GCN. GraphRNA \cite{2019GraphRNA} and ARWMF \cite{2019ARWMF} are representatives of attributed random walk-based methods, which apply random walks on the node-attribute bipartite graphs and can potentially tackle the attribute-missing issue. Since attribute-missing restoration is similar to cold-start recommendation problem, thus one representative cold-start recommendation method called Hers \cite{2019HERS} is involved as a baseline. Further, we report the performance of two attribute-incomplete methods, i.e, GINN \cite{2020GINN}, GCNMF \cite{2021GCNMF} and a state-of-the-art attribute-missing method, i.e., SAT \cite{2020SAT}. Table \ref{III} and Table \ref{IV} summarize the performance comparison on the tasks of profiling and node classification.

On profiling task, four observations can be obtained from Table \ref{III}:
1) SAGA shows the best performance in terms of six metrics against all compared baselines over four datasets. For instance, SAT has been considered as the strongest attribute-missing GRL framework, our method exceeds it by 1.91\%/2.60\%, 2.97\%/4.11\%, 1.37\%/2.20\%, and 3.06\%/0.56\% in terms of Recall@50 and NDCG@50 on four datasets, which verifies the effectiveness of structure-attribute mutual enhanced learning strategy in handling attribute-missing graphs; 2) it can be seen that SAGA consistently outperforms the attributed rand walk-based methods. This is because the operation of random walks may introduce noise to the learning process, which affects the quality of restored attributes; 3) our SAGA achieves better performance than GCN, GraphSage, and GAT, all of which have been demonstrated strong representation learning capability on complete graphs, while these methods are unable to effectively handle attribute-missing cases; 4) since NeighAggre and VAE isolate the processes of data completion and network learning, thus both obtain unpromising results. 

On node classification task, as reported in Table \ref{IV}, we can find that 1)
our method achieves the best average performance in terms of accuracy on five of six datasets. For instance, SAGA exceeds SAT by 1.86\%, 3.26\%, 3.46\%, 0.73\%, and 5.18\%, 3.14\% accuracy increment. These results well demonstrate that our method could learn a more discriminative latent embedding for data completion by taking full advantages of attribute and structure information, thus boosting the performance of downstream tasks; 2) compared with GINN and GCNMF, our method gains 14.83\%, 5.85\%, 6.93\%, 4.57\%, 18.55\%, and 1.37\% accuracy increment, which indicates that our method is competent to handle attribute-missing graphs, while incomplete GRL methods can not provide effective solutions; 3) note that GraphRNA and ARWME achieve slightly better results compared with ours on Pubmed dataset, this is because GraphRNA and ARWME could naturally model the correlation between attribute dimension, especially for handling real-valued data.

Overall, the results of both profiling and node classification tasks have demonstrated the superiority of SAGA in solving attribute-missing graph representation learning.

\subsection{Ablation Comparison}
\begin{figure}[!t]
\centering
\includegraphics[width=3.2in]{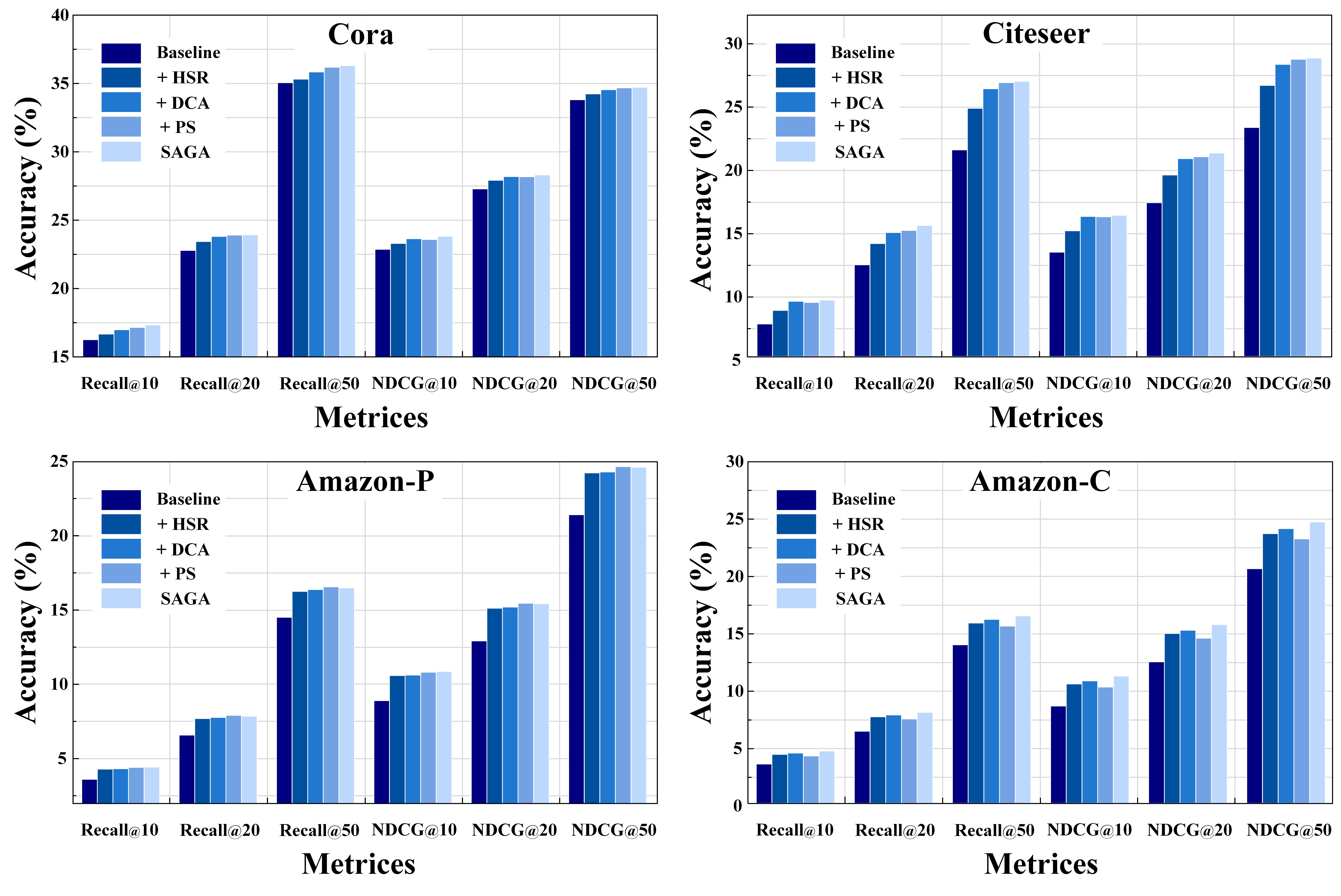}
\caption{Ablation study on four datasets. The baseline is a naive graph auto-encoder. +HSR and +DCA indicate that the baseline adopts HSR module and DCA module, respectively. +PS denotes a pseudo-siamese counterpart of SAGA.}
\label{3}
\end{figure}

In this section, we design an ablation study to clearly demonstrate the effectiveness of the proposed SAGA. A naive graph auto-encoder is developed as the baseline. +HSR and +DCA indicate that the baseline adopts HSR module and DCA module, respectively. +PS denotes a pseudo-siamese counterpart of SAGA. We experimentally compare all methods on four datasets and report the results in Fig. \ref{3}. We can see that 1) +HSR method and +DCA method consistently improve the baseline in terms of all metrics over all datasets. Taking the results on Amazon-C for example, +HSR method and +DCA method gain 2.12\%/2.21\% and 3.43\%/3.49\% accuracy increment in terms of Recall@50/NDCG@50. These results verify the effectiveness of sufficient interaction between attribute and structure information (i.e., information filtering and graph structure refinement) for data completion. We can obtain similar observations from the results on other metrics and datasets; 2) this ablation study also reveals the advantage of our siamese architecture, which can help to better exploit the two-source information for data imputation.
As seen, the proposed SAGA demonstrates slightly better performance than the pseudo-siamese counterpart. According to these observations, the effectiveness of the proposed components in SAGA has been clearly verified.

\begin{figure}[!t]
\centering
\includegraphics[width=3.0in]{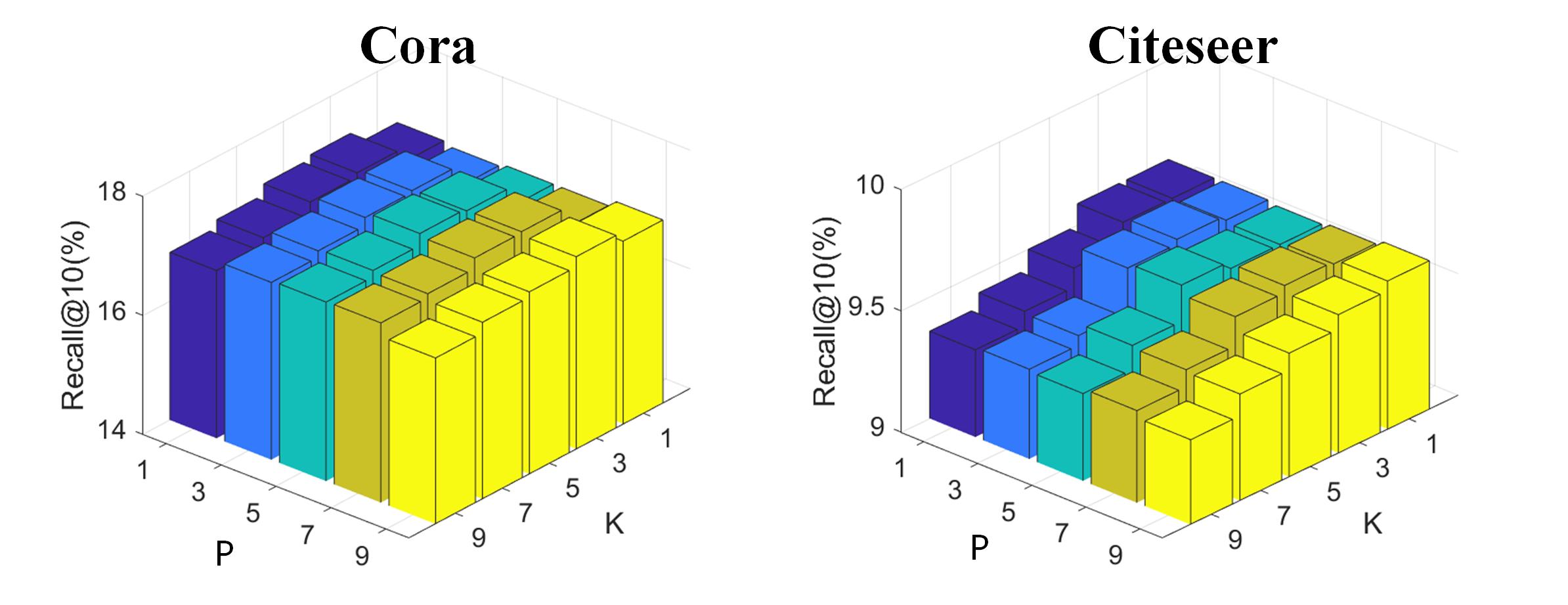}
\caption{Network performance with variations of $P$ and $K$.}
\label{4}
\end{figure}

\subsection{Analysis of Hyper-parameters}
\noindent{\textbf{Parameter Analysis of $P$ and $K$}}. In this part, the effect of two hyper-parameters has been investigated on Cora and Citeseer. In Fig. \ref{4}, the performance variation of SAGA when $P$ and $K$ vary from 1 to 9 is presented, where we can see that 1) for a fixed $P$, $K$ value between 3 and 5 achieves better performance; 2) for a certain $K$, SAGA obtains stable results when $P$ varies from 3 to 7; 3) SAGA tends to perform well by setting $P$ and $K$ to 5 across both datasets.
 
\begin{figure}[!t]
\centering
\includegraphics[width=3.2in]{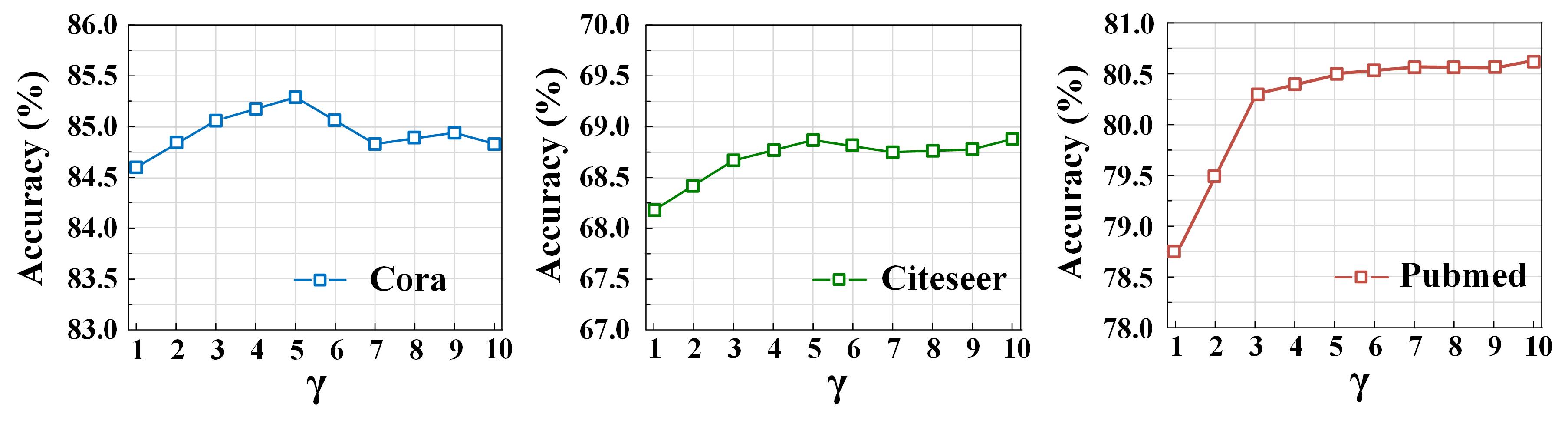}
\caption{Classification accuracy vs. hyper-parameter $\gamma$.
}
\label{5}
\end{figure}

\begin{figure}[!t]
\centering
\includegraphics[width=3.0in]{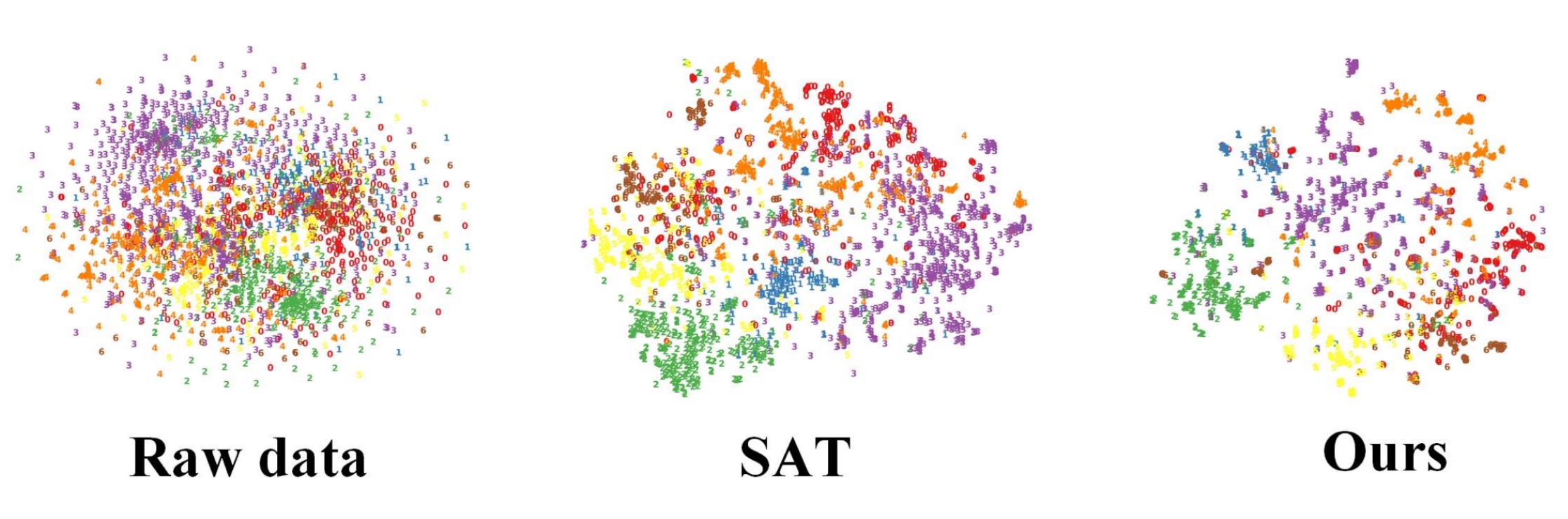}
\caption{\textit{t}-SNE visualization comparisons on Cora.}
\label{6}
\end{figure}
\noindent{\textbf{Parameter Analysis of $\gamma$}}.
Furthermore, we conduct experiments to show the effect of hyper-parameter $\gamma$ in Eq.(9) on three datasets. Fig. \ref{5} presents the results of node classification with different $\gamma$. We can observe that 1) $\gamma$ is effective in improving the performance; 2) the ACC metric increases to a higher value when $\gamma$ varies from 0 to 5; 3) with the increasing value of $\gamma$, the performance on Cora tends to drop but still keeps better than that of the baseline ($\gamma$ = 1); 4) SAGA performs well by setting $\gamma$ to 5 across all datasets.

\subsection{\textit{t}-SNE Visualization Comparison}
To intuitively verify the effectiveness of SAGA, we visualize the distribution of the learned graph embedding in two-dimensional space by employing \textit{t}-SNE algorithm~\cite{2008Visualizing}. As illustrated in Fig. \ref{6}, the visual results show that our proposed SAGA presents a cleaner division between classes compared with SAT. 

More experimental results are presented in Appendix B.

\section{Conclusion}
In this paper, we propose the SAGA method to handle attribute-missing graphs. In our network, two core components, i.e., DCA and HSR modules, take full advantages of structures and attributes, and allow both types of information to sufficiently interact with each other in a siamese framework. In this way, unreliable information is filtered while more informative information can be well collected and preserved, which effectively enhances the discriminative capacity of the learned latent embedding for data completion. Moreover, our siamese design is able to assist in boosting the learning capability of SAGA. Extensive experiments on six benchmark datasets have demonstrated the effectiveness and superiority of the proposed method. Future work may aim to extend SAGA to handle multi-view attribute-missing GRL, where various observed information from different views can be exploited to further improve the performance.



\bibliography{myAAAIBib}
\bibliographystyle{aaai}

\end{document}